\pdfoutput=1

\documentclass[11pt]{article}

\usepackage{acl2023} 
\usepackage{times}
\usepackage{latexsym}
\usepackage{booktabs}
\usepackage{multirow}
\usepackage{enumitem}
\usepackage[T1]{fontenc}

\usepackage[utf8]{inputenc}

\usepackage{microtype}

\usepackage{inconsolata}

\title{Overview of the Problem List Summarization (ProbSum) 2023 Shared Task on Summarizing Patients' Active Diagnoses and Problems from Electronic Health Record Progress Notes}

\author{Yanjun Gao$^{1}$, Dmitriy Dligach$^{2}$, Timothy Miller$^{3}$, \\
\textbf{Matthew M. Churpek}$^{1}$ and \textbf{Majid Afshar}$^{1}$ \\
$^{1}$University of Wisconsin,
$^{2}$Loyola University Chicago, \\ 
$^{3}$Boston Children's Hospital and Harvard Medical School \\
\texttt{$^{1}$\{ygao, mchurpek, mafshar\} @medicine.wisc.edu, }  \\ \texttt{$^{2}$ddligach@luc.edu,$^{3}$Timothy.Miller@childrens.harvard.edu}\\}

\begin{document}
\maketitle
\begin{abstract}
The BioNLP Workshop 2023 initiated the launch of a shared task on Problem List Summarization (ProbSum) in January 2023. The aim of this shared task is to attract future research efforts in building NLP models for real-world diagnostic decision support applications, where a system generating relevant and accurate diagnoses will augment the healthcare providers’ decision-making process and improve the quality of care for patients. The goal for participants is to develop models that generated a list of diagnoses and problems using input from the daily care notes collected from the hospitalization of critically ill patients. Eight teams submitted their final systems to the shared task leaderboard. In this paper, we describe the tasks, datasets, evaluation metrics, and baseline systems. Additionally, the techniques and results of the evaluation of the different approaches tried by the participating teams are summarized.  

\end{abstract}

\section{Introduction}
Electronic health records (EHR) document patients' daily progress and, therefore, are an essential component of hospital care~\citep{brown2014physicians}. However, note-taking in the EHR, while necessary, suffers from note bloat and information overload, causing challenges for end-users~\citep{liu2022note}. The cognitive burden on providers using EHR documents remains high and only increases in our digital era~\citep{furlow2020information, hultman2019challenges}. Summarization tasks in clinical natural language processing (NLP) are a promising approach to overcome note bloat and help extract relevant, active diagnoses to help mitigate diagnostic errors. Augmented intelligence with computerized diagnostic decision support is a potential solution to improve care throughput and quality at the bedside~\citep{demner2009can, lederman2022tasks}.  

Over the past two decades, many tasks in the field of clinical NLP (cNLP) have focused on information extraction, document classification (e.g., named entity recognition), and relation extraction, as noted in a recent scoping review of cNLP tasks~\citep{gao2022scoping}. As generative NLP systems have advanced significantly in recent years, the field has a growing opportunity on taking advantage of the latest advancements in generative large language models such as T5~\cite{raffel2020exploring}, Galactica~\cite{taylorgalactica}, and GPT~\cite{floridi2020gpt}. Tasks of clinical text generation include radiology report generation~\cite{xue2018multimodal,ben-abacha-etal-2021-overview}, clinical note generation~\cite{yang2020generating,krishna-etal-2021-generating,grambow2022domain}, and patients' diagnoses and problems summarization~\cite{gao2022summarizing}. For the BioNLP Workshop 2023, the three shared tasks for the BioNLP Workshop 2023 all focused on biomedical text generation, in line with the current research trend on generative NLP models. This paper presents the overview of the shared task for the BioNLP Workshop 2023, \textit{Problem List Summarization}. 

The Problem List Summarization (\textbf{ProbSum}) task, initially proposed by ~\citet{gao2022summarizing}, aimed to facilitate the development of NLP models for downstream applications in diagnostic decision support systems. The goal of this shared task was to attract future research efforts in developing NLP systems for real-world decision support applications, where a system generating relevant and accurate diagnoses can augment the healthcare providers’ decision-making process.

The ProbSum task utilized the daily progress notes, a documentation approach that is ubiquitous in the EHR with a specific format that is taught in medical schools. The progress note contains the following four major sections: Subjective, Objective, Assessment, and Plan (SOAP)~\cite{weed1969medical}. Participants developed models to create a list of relevant diagnoses or problems based on the information present in the Subjective, Objective, and Assessment sections of the note. The Plan contained the diagnoses and problems that served as the ground truth labels.

Summarizing the list of problems and diagnoses from daily progress notes is a challenging task for several reasons. First, progress notes may contain old diagnoses that are no longer relevant or list diagnoses that are not described or mentioned elsewhere in the note.  Second, the progress note may contain extra information needed for billing purposes or quality improvement metrics that do not contain diagnoses. Further, the data collected in the progress note may be embedded from other parts of the EHR, such as laboratory results or medications. The structured text and tabular data alongside unstructured text present a multimodal problem. Lastly, progress notes may contain medical jargon and abbreviations (e.g. ``GI'', ``EGD'' in Figure~\ref{fig:task_example}) that can be challenging for NLP models to understand and summarize. To accurately summarize patient problems, models must possess a deep understanding of the medical concepts and terminology involved, as well as the broader context of the patient's medical history and current condition. 


\section{Task Description}
Healthcare providers, including physicians, write daily progress notes to record the patient's medical status, treatment, and progress while they are hospitalized. These notes contain essential information, such as vital signs, medications, procedures, and any significant developments in the patient's condition. Daily progress notes are crucial for monitoring the patient's overall health and ensuring that the healthcare team is informed of any changes or updates in their treatment. 

Daily progress notes are formatted using the SOAP Format. The \textit{Subjective} section of a SOAP format daily progress note comprises the patient's self-reported symptoms, concerns, and medical history. The \textit{Objective} section consists of objective data collected by healthcare providers during observation or examination, such as vital signs (e.g., blood pressure, heart rate), laboratory results, or physical exam findings. The \textit{Assessment} section summarizes the patient's overall condition with a focus on the most active problems/diagnoses for that day. Finally, the \textit{Plan} section contains multiple subsections, each outlining a diagnosis/problem and its treatment plan. The aim of the task was to predict the list of problems and diagnoses outlined in the \textit{Plan} section~\cite{weed1969medical}. Figure~\ref{fig:task_example} presents an input example from a progress note with Subjective, Objective, and Assessment sections, and the Ground Truth problem/diagnosis List annotated from the Plan section.   

\begin{figure}
    \centering
    \small 
    \begin{tabular}{l} \toprule 
         \textbf{(Subjective)} \\
         Chief Complaint: CC: Diarrhea  Admission: GI Bleed 24  \\
         Hour Events: - GI performed EGD: diffuse erythema/\\
         ulceration, in esophagus, stomach, duodenum.\ldots  \\ \midrule 
         \textbf{(Objective)} \\ 
         07:39 AM Vital signs Hemodynamic monitoring Fluid \\ balance 24 hours Since  AM Tmax: 38.1 C (100.6  \\
         Tcurrent: 37.7 C (99.8 HR: 122 (96 - 128) bpm \ldots \\
        \midrule 
         \textbf{(Assessment)} \\ 
         72yo M with h/o metastatic rectal CA and SVC syndrome\\ 
         on lovenox admitted with intractable diarrhea thought \\
         chemo now with BRBPR and hematemesis. \\ 
         \midrule 
         \textbf{(Plan Subsections with Problem List Annotation)} \\ 
         \# \colorbox{green!20}{GI Bleed/diarrhea}:
         Given EGD results, his bleed \\ 
         appears to be from 
         \colorbox{green!20}{diffuse ulceration from the esophagus} \\ 
         \colorbox{green!20}{through the duodenum.} As diarrhea has been \\
         mucuous with pink tinge, \ldots \\  
         \midrule
         \textbf{(Ground Truth Problem List)} \\
         GI Bleed; diarrhea; diffuse ulceration from the esophagus \\  through the duodenum; Acute Renal Failure; SVC  \\  Syndrome; Rectal CA; \\ 
         \bottomrule  
    \end{tabular}
    \vspace{-.1in}
    \caption{An example progress note with Subjective, Objective, Assessment sections and the Ground Truth Summary annotated from Plan sections. Annotated text spans are highlighted in \colorbox{green!20}{color box}. In the ground truth, annotated diagnosis/problem are concatenated by semicolons. We use $\ldots$ to denote the continuation of the notes due to space constraints.}
    \label{fig:task_example}
\end{figure}

\section{Data Description}
The task contained 768 hospital daily progress notes and 2783 diagnoses in the training set, which was previously described in more detail~\cite{gao2022summarizing}. A critical care physician and clinical informatics expert annotated a test set of 237 daily progress notes with only one progress note per patient to avoid redundancy. 

\subsection{Data source}
The progress notes were sourced from MIMIC-III, a publicly available dataset of de-identified EHR data from approximately 60,000 hospital critical care admissions at Beth Israel Deaconess Medical Center in Boston, Massachusetts~\cite{johnson2016mimic}\footnote{Data Use Agreement is required for all participants}. The progress note types from MIMIC-III included a total of 84 progress note types across medical, surgical, cardiovascular, trauma, and neurology specialties. Other note types were excluded such as Nursing Progress Notes and Social Worker Progress Notes because they were not structured in the SOAP format. 

\subsection{Annotation} 
The corpora for Problem List Summarization followed the annotation framework and guidelines previously described in~\cite{gao2022hierarchical}. The annotation was performed in the INCEpTION annotation platform~\cite{tubiblio106270}. 
The goal of the annotation was to label lists of relevant problems/diagnoses from the Plan subsections. The annotators first marked the text span of the Assessment and Plan subsections. For each Plan subsection, the annotators marked the text span for the Problem, separating the diagnosis/problems from the treatment or action plans. Since the annotation was conducted on individual progress notes, the guidelines specified that only the active and pertinent diagnoses should be considered, taking into account the patient's condition as described in those specific progress notes. Consequently, only the active and relevant diagnoses/problems were included in this shared task. 

In the original dataset introduced by~\citet{gao2022tasks}, 
there were 768 annotated progress notes for training and test set. We merged the original split into training data for this shared task, and had a critical care physician annotate another 237 progress notes for a new test set. 
We further parsed the annotation into CSV files that had five columns: FILE ID, Subjective Sections, Objective Sections, Assessment, and Summary (Ground Truth). The FILE ID corresponded to the unique IDs of clinical notes in the original MIMIC-III dataset. 
Participants could choose a combination of Subjective, Objective and Assessment as input. They were also free to incorporate any other domains of the MIMIC data that were related to the patient's progress note. 

For the final test set, we found the top 10 most diagnoses are: Anemia, auricular fibrillation, acute renal failure, Septicemia, Sepsis,  Respiratory failure, Hypertension, Congestive heart failure, thrombocytopenia and Pneumonia. 

\section{Evaluation}
As the evaluation metric, we employed ROUGE-L, a measurement that captures the Longest Common Subsequence (LCS) between a generated summary and its corresponding reference summary~\citep{lin2004rouge}. In what follows, we report ROUGE-L Precision (RL-P),  ROUGE-L Recall (RL-R), and ROUGE-L F-score (RL-F). Although we acknowledge the limitations of using ROUGE as an evaluation approach, as it does not assess semantics, the field of automated evaluation for clinical note summarization is still in its early stages, with most research relying on human evaluation besides conventional metrics like ROUGE~\cite{gao2022summarizing,otmakhova2022patient}. Therefore, for the purpose of the shared task, which requires ranking systems, we have chosen to use ROUGE as our evaluation metric.

\section{Participation}
We released the training data and test data through PhysioNet\footnote{https://physionet.org/}, which serve as the data owners of the MIMIC-III data and the executors of the Data Use Agreement with participants. During System Evaluation Phase, we created a CodaLab competition to manage the system submission and leaderboard~\cite{codalab_competitions}. To be specific, the CodaLab was configured to receive system prediction outputs in a text file format, and subsequently ran the ROUGE-L evaluation script to produce scores. Each user was permitted to submit a maximum of 30 runs. At the time of writing this paper, eight teams from the United Kingdom, India, Spain, South Korea, Ireland, Australia, and Mexico  submitted a total of 164 runs. Along with the system output text file, we asked participants to provide a brief description of their methodology; however, some users submitted their results without providing us with any system details. 

\section{Results}
\subsection{Baseline}
According to the findings of \citeauthor{gao2022summarizing}(\citeyear{gao2022summarizing}), the ProbSum task's baseline result was established by applying pre-trained sequence-to-sequence language models, specifically T5~\cite{raffel2020exploring} and BART~\cite{lewis2020bart}. T5 produced the most favorable set of results. To achieve the best ROUGE-L F-score of 18.80, T5 was subjected to domain adaptive pre-training~\cite{gururangan2020don} and concept masking, which involved randomly masking the concepts identified by the concept extractor and continually training T5 on MIMIC data. \citet{gao2022summarizing}~further found that T5 produced the best results when Subjective and Assessment sections were used as inputs. However, inputting Objective sections could result in noisy output. 

\subsection{Main results}

\begin{table}[ht]
\small
    \centering
    \begin{tabular}{l|l|l|l} \toprule
       \textbf{ Team} & \textbf{RL-P} & \textbf{RL-R} &\textbf{RL-F}  \\ \midrule
       CUED  & 41.69  & 30.51& 32.77  \\ \midrule 
       PULSAR  & 44.30 & 27.18 & 31.14 \\ \midrule 
       Pune Institute of Comp. Tech  & 41.39 & 22.96 & 27.44 \\ \midrule 
       Universitat Politècnica & \multirow{2}{*}{32.86} & \multirow{2}{*}{24.79} & \multirow{2}{*}{25.92} \\
       de Catalunya  &  & &  \\ \midrule  
       Sungkyunkwan Univ.  & 34.44 & 19.55 & 22.39 \\ \midrule  
       Deakin Univ.  & 28.66 & 19.11 & 20.84 \\ \midrule  
       Univ. of Limerick  & 39.44 & 15.18 & 19.67 \\ \midrule  
       Universidad Nacional & \multirow{2}{*}{11.88} & \multirow{2}{*}{13.37} & \multirow{2}{*}{11.83}\\
       Autónoma De México  &  & &  \\  
       \bottomrule
    \end{tabular}
    \vspace{-.1in}
    \caption{Best-performing submission results from the eight participation teams. We report ROUGE-L Precision (RL-P), Recall (RL-R) and F-score (RL-F). }
    \label{tab:main_results}
\end{table}

Table~\ref{tab:main_results} displays the best results submitted by eight participating teams, with Team CUED (University of Cambridge) achieving the highest F1 score of 32.77. Team PULSAR (University of Manchester) closely followed with a score of 31.44. The team from Pune Institute of Computer Technology held the third spot with an F1 score of 27.44. The F1 score reported by Team Universitat Politècnica de Catalunya was 25.92, while Team Sungkyunkwan University and Team Deakin University reported F1 scores of 22.39 and 20.84, respectively. Team University of Limerick reported an F1 score of 19.67, which is near the baseline. Team Universidad Nacional Autónoma De México reported the lowest F1 score of 11.83. 

\subsection{Methods overview}

We asked participants to submit a post-participation survey that further described their approaches, and received four teams' response. The survey contained questions regarding the best-performing systems and input setting, as listed in Figure~\ref{fig:survey}. Four of the eight teams submitted their answers to the survey questions: team CUED, PULSAR, Universitat Politècnica de Catalunya and Univ. of Limerick.  
\begin{figure}[ht]
    \centering
\small 
\begin{itemize}[noitemsep]
    \item Method descriptions
    \item What is the input to your best performing system? (Options: Subjective, Objective, Assessment)
    \item Did you use language model?
    \item Did you use medical doctors’ expertise? (One of your team member is medical doctor, or your team has consulted with medical doctor) 
    \item Did your model use other resources (e.g. during pre-training) besides MIMIC? (E.g. PubMed text, UMLS) 
\end{itemize}
\vspace{-.1in}
    \caption{Post-participation Survey Questions}
    \label{fig:survey}
\end{figure}

The collected surveys revealed that all teams were utilizing transformer based language models, like T5 and BERT~\cite{kenton2019bert}. T5 checkpoints trained on medical resources were popular choices for methods, such as Clinical-T5~\cite{lehmanclinical}. The medical resources employed for training, apart from MIMIC, included Unified Medical Language System (UMLS)~\cite{bodenreider2004unified}, BC5CDR (BioCreative V CDR corpus)~\cite{li2016biocreative}, and i2b2 2010 dataset~\cite{uzuner20112010}. All teams employed the Subjective and Assessment sections as inputs, while two teams used Objective sections as additional inputs. There were no teams that included a member who was a medical provider or had consulted with medical experts.  

Team CUED achieved the best performance by using an ensembled clinical T5 model. On the other hand, Team PULSAR used Flan-T5~\cite{chung2022scaling} and GPT2XL~\cite{radford2019language}.  Team Universitat Politècnica de Catalunya trained a BERT-based Named Entity Recognition (NER) BIO tagger (NER) system to extract keywords from input and composed the output. Finally, team University of Limerick developed a hybrid summarization system that utilized the Pagerank algorithm~\cite{langville2006google} and QuickUMLS~\cite{soldaini2016quickumls} on a concept graph to extract the most important sentences. They further used a fine-tuned T5 model with concept masking to generate an abstractive summary of the generated summary.

\section{Discussion}

The 2023 ProbSumm BioNLP Workshop revealed that transformer-based language models trained on medical resources are capable of generating summaries from medical texts.


While this finding was consistent with previous research that showed the effectiveness of these models in various NLP tasks, including text summarization, the ProbSumm task proved to be very difficult and is far from solved with a best F1 score of 32.77.

None of the teams involved in the task included a medical provider or sought consultation from medical experts while developing their models. This could be attributed to practical limitations such as time and resource constraints. However, incorporating the expertise of medical professionals during the development process may provide valuable insights into the clinical implications of the generated summaries, resulting in the creation of more clinically relevant and useful models. Given the effectiveness of incorporating human feedback in various NLP tasks~\cite{ouyang2022training}, we recommend future research to explore the performance of involving medical experts in the development and evaluation through a human-in-the-loop approach. 

\section{Conclusion}
The ProbSum task, part of the BioNLP Shared Task 1A, focused on summarizing patients' diagnoses from daily progress notes. Eight teams from 7 countries submitted the final systems for evaluation, and the top-performing system has achieved substantial gains from the baseline with a new state-of-the-art result of 32.77. 
The experience gained by participants can inform the development of more effective strategies for medical text summarization, which could have significant implications for augmenting computerized diagnostic decision support systems at the bedside and potentially translate into better patient care.

\bibliography{reference}

\begin{thebibliography}{36}
\expandafter\ifx\csname natexlab\endcsname\relax\def\natexlab#1{#1}\fi

\bibitem[{Ben~Abacha et~al.(2021)Ben~Abacha, Mrabet, Zhang, Shivade, Langlotz,
  and Demner-Fushman}]{ben-abacha-etal-2021-overview}
Asma Ben~Abacha, Yassine Mrabet, Yuhao Zhang, Chaitanya Shivade, Curtis
  Langlotz, and Dina Demner-Fushman. 2021.
\newblock \href {https://doi.org/10.18653/v1/2021.bionlp-1.8} {Overview of the
  {MEDIQA} 2021 shared task on summarization in the medical domain}.
\newblock In \emph{Proceedings of the 20th Workshop on Biomedical Language
  Processing}, pages 74--85, Online. Association for Computational Linguistics.

\bibitem[{Bodenreider(2004)}]{bodenreider2004unified}
Olivier Bodenreider. 2004.
\newblock The unified medical language system (umls): integrating biomedical
  terminology.
\newblock \emph{Nucleic acids research}, 32(suppl\_1):D267--D270.

\bibitem[{Brown et~al.(2014)Brown, Marquard, Amster, Romoser, Friderici, Goff,
  and Fisher}]{brown2014physicians}
PJ~Brown, JL~Marquard, B~Amster, M~Romoser, J~Friderici, S~Goff, and D~Fisher.
  2014.
\newblock What do physicians read (and ignore) in electronic progress notes?
\newblock \emph{Applied clinical informatics}, 5(02):430--444.

\bibitem[{Chung et~al.(2022)Chung, Hou, Longpre, Zoph, Tay, Fedus, Li, Wang,
  Dehghani, Brahma et~al.}]{chung2022scaling}
Hyung~Won Chung, Le~Hou, Shayne Longpre, Barret Zoph, Yi~Tay, William Fedus,
  Eric Li, Xuezhi Wang, Mostafa Dehghani, Siddhartha Brahma, et~al. 2022.
\newblock Scaling instruction-finetuned language models.
\newblock \emph{arXiv preprint arXiv:2210.11416}.

\bibitem[{Demner-Fushman et~al.(2009)Demner-Fushman, Chapman, and
  McDonald}]{demner2009can}
Dina Demner-Fushman, Wendy~W Chapman, and Clement~J McDonald. 2009.
\newblock What can natural language processing do for clinical decision
  support?
\newblock \emph{Journal of biomedical informatics}, 42(5):760--772.

\bibitem[{Devlin et~al.(2019)Devlin, Chang, Lee, and
  Toutanova}]{kenton2019bert}
Jacob Devlin, Ming-Wei Chang, Kenton Lee, and Lee~Kristina Toutanova. 2019.
\newblock Bert: Pre-training of deep bidirectional transformers for language
  understanding.
\newblock In \emph{Proceedings of NAACL-HLT}, pages 4171--4186.

\bibitem[{Floridi and Chiriatti(2020)}]{floridi2020gpt}
Luciano Floridi and Massimo Chiriatti. 2020.
\newblock Gpt-3: Its nature, scope, limits, and consequences.
\newblock \emph{Minds and Machines}, 30:681--694.

\bibitem[{Furlow(2020)}]{furlow2020information}
Bryant Furlow. 2020.
\newblock Information overload and unsustainable workloads in the era of
  electronic health records.
\newblock \emph{The Lancet Respiratory Medicine}, 8(3):243--244.

\bibitem[{Gao et~al.(2022{\natexlab{a}})Gao, Caskey, Miller, Sharma, Churpek,
  Dligach, and Afshar}]{gao2022tasks}
Y.~Gao, J.~Caskey, T.~Miller, B.~Sharma, M.~Churpek, D.~Dligach, and M.~Afshar.
  2022{\natexlab{a}}.
\newblock \href {https://doi.org/10.13026/wks0-w041} {{Tasks 1 and 3 from
  Progress Note Understanding Suite of Tasks: SOAP Note Tagging and Problem
  List Summarization}}.
\newblock {PhysioNet}.

\bibitem[{Gao et~al.(2022{\natexlab{b}})Gao, Dligach, Christensen, Tesch,
  Laffin, Xu, Miller, Uzuner, Churpek, and Afshar}]{gao2022scoping}
Yanjun Gao, Dmitriy Dligach, Leslie Christensen, Samuel Tesch, Ryan Laffin,
  Dongfang Xu, Timothy Miller, Ozlem Uzuner, Matthew~M Churpek, and Majid
  Afshar. 2022{\natexlab{b}}.
\newblock A scoping review of publicly available language tasks in clinical
  natural language processing.
\newblock \emph{Journal of the American Medical Informatics Association},
  29(10):1797--1806.

\bibitem[{Gao et~al.(2022{\natexlab{c}})Gao, Dligach, Miller, Tesch, Laffin,
  Churpek, and Afshar}]{gao2022hierarchical}
Yanjun Gao, Dmitriy Dligach, Timothy Miller, Samuel Tesch, Ryan Laffin,
  Matthew~M Churpek, and Majid Afshar. 2022{\natexlab{c}}.
\newblock Hierarchical annotation for building a suite of clinical natural
  language processing tasks: Progress note understanding.
\newblock In \emph{LREC... International Conference on Language Resources \&
  Evaluation:[proceedings]. International Conference on Language Resources \&
  Evaluation}, volume 2022, page 5484. NIH Public Access.

\bibitem[{Gao et~al.(2022{\natexlab{d}})Gao, Dligach, Miller, Xu, Churpek, and
  Afshar}]{gao2022summarizing}
Yanjun Gao, Dmitriy Dligach, Timothy Miller, Dongfang Xu, Matthew~MM Churpek,
  and Majid Afshar. 2022{\natexlab{d}}.
\newblock Summarizing patients’ problems from hospital progress notes using
  pre-trained sequence-to-sequence models.
\newblock In \emph{Proceedings of the 29th International Conference on
  Computational Linguistics}, pages 2979--2991.

\bibitem[{Grambow et~al.(2022)Grambow, Zhang, and Schaaf}]{grambow2022domain}
Colin Grambow, Longxiang Zhang, and Thomas Schaaf. 2022.
\newblock In-domain pre-training improves clinical note generation from
  doctor-patient conversations.
\newblock In \emph{Proceedings of the First Workshop on Natural Language
  Generation in Healthcare}, pages 9--22.

\bibitem[{Gururangan et~al.(2020)Gururangan, Marasovi{\'c}, Swayamdipta, Lo,
  Beltagy, Downey, and Smith}]{gururangan2020don}
Suchin Gururangan, Ana Marasovi{\'c}, Swabha Swayamdipta, Kyle Lo, Iz~Beltagy,
  Doug Downey, and Noah~A Smith. 2020.
\newblock Don’t stop pretraining: Adapt language models to domains and tasks.
\newblock In \emph{Proceedings of the 58th Annual Meeting of the Association
  for Computational Linguistics}, pages 8342--8360.

\bibitem[{Hultman et~al.(2019)Hultman, Marquard, Lindemann, Arsoniadis,
  Pakhomov, and Melton}]{hultman2019challenges}
Gretchen~M Hultman, Jenna~L Marquard, Elizabeth Lindemann, Elliot Arsoniadis,
  Serguei Pakhomov, and Genevieve~B Melton. 2019.
\newblock Challenges and opportunities to improve the clinician experience
  reviewing electronic progress notes.
\newblock \emph{Applied clinical informatics}, 10(03):446--453.

\bibitem[{Johnson et~al.(2016)Johnson, Pollard, Shen, Lehman, Feng, Ghassemi,
  Moody, Szolovits, Anthony~Celi, and Mark}]{johnson2016mimic}
Alistair~EW Johnson, Tom~J Pollard, Lu~Shen, Li-wei~H Lehman, Mengling Feng,
  Mohammad Ghassemi, Benjamin Moody, Peter Szolovits, Leo Anthony~Celi, and
  Roger~G Mark. 2016.
\newblock Mimic-iii, a freely accessible critical care database.
\newblock \emph{Scientific data}, 3(1):1--9.

\bibitem[{Klie et~al.(2018)Klie, Bugert, Boullosa, de~Castilho, and
  Gurevych}]{tubiblio106270}
Jan-Christoph Klie, Michael Bugert, Beto Boullosa, Richard~Eckart de~Castilho,
  and Iryna Gurevych. 2018.
\newblock \href {http://tubiblio.ulb.tu-darmstadt.de/106270/} {The inception
  platform: Machine-assisted and knowledge-oriented interactive annotation}.
\newblock In \emph{Proceedings of the 27th International Conference on
  Computational Linguistics: System Demonstrations}, pages 5--9. Association
  for Computational Linguistics.
\newblock Event Title: The 27th International Conference on Computational
  Linguistics (COLING 2018).

\bibitem[{Krishna et~al.(2021)Krishna, Khosla, Bigham, and
  Lipton}]{krishna-etal-2021-generating}
Kundan Krishna, Sopan Khosla, Jeffrey Bigham, and Zachary~C. Lipton. 2021.
\newblock \href {https://doi.org/10.18653/v1/2021.acl-long.384} {Generating
  {SOAP} notes from doctor-patient conversations using modular summarization
  techniques}.
\newblock In \emph{Proceedings of the 59th Annual Meeting of the Association
  for Computational Linguistics and the 11th International Joint Conference on
  Natural Language Processing (Volume 1: Long Papers)}, pages 4958--4972,
  Online. Association for Computational Linguistics.

\bibitem[{Langville and Meyer(2006)}]{langville2006google}
Amy~N Langville and Carl~D Meyer. 2006.
\newblock \emph{Google's PageRank and beyond: The science of search engine
  rankings}.
\newblock Princeton university press.

\bibitem[{Lederman et~al.(2022)Lederman, Lederman, and
  Verspoor}]{lederman2022tasks}
Asher Lederman, Reeva Lederman, and Karin Verspoor. 2022.
\newblock Tasks as needs: reframing the paradigm of clinical natural language
  processing research for real-world decision support.
\newblock \emph{Journal of the American Medical Informatics Association},
  29(10):1810--1817.

\bibitem[{Lehman et~al.(2023)Lehman, Hernandez, Mahajan, Wulff, Smith, Ziegler,
  Nadler, Szolovits, Johnson, and Alsentzer}]{lehmanclinical}
Eric Lehman, Evan Hernandez, Diwakar Mahajan, Jonas Wulff, Micah~J Smith,
  Zachary Ziegler, Daniel Nadler, Peter Szolovits, Alistair Johnson, and Emily
  Alsentzer. 2023.
\newblock Do we still need clinical language models?
\newblock \emph{arXiv preprint arXiv:2302.08091}.

\bibitem[{Lewis et~al.(2020)Lewis, Liu, Goyal, Ghazvininejad, Mohamed, Levy,
  Stoyanov, and Zettlemoyer}]{lewis2020bart}
Mike Lewis, Yinhan Liu, Naman Goyal, Marjan Ghazvininejad, Abdelrahman Mohamed,
  Omer Levy, Veselin Stoyanov, and Luke Zettlemoyer. 2020.
\newblock Bart: Denoising sequence-to-sequence pre-training for natural
  language generation, translation, and comprehension.
\newblock In \emph{Proceedings of the 58th Annual Meeting of the Association
  for Computational Linguistics}, pages 7871--7880.

\bibitem[{Li et~al.(2016)Li, Sun, Johnson, Sciaky, Wei, Leaman, Davis,
  Mattingly, Wiegers, and Lu}]{li2016biocreative}
Jiao Li, Yueping Sun, Robin~J Johnson, Daniela Sciaky, Chih-Hsuan Wei, Robert
  Leaman, Allan~Peter Davis, Carolyn~J Mattingly, Thomas~C Wiegers, and Zhiyong
  Lu. 2016.
\newblock Biocreative v cdr task corpus: a resource for chemical disease
  relation extraction.
\newblock \emph{Database}, 2016.

\bibitem[{Lin(2004)}]{lin2004rouge}
Chin-Yew Lin. 2004.
\newblock Rouge: A package for automatic evaluation of summaries.
\newblock In \emph{Text summarization branches out}, pages 74--81.

\bibitem[{Liu et~al.(2022)Liu, Capurro, Nguyen, and Verspoor}]{liu2022note}
Jinghui Liu, Daniel Capurro, Anthony Nguyen, and Karin Verspoor. 2022.
\newblock “note bloat” impacts deep learning-based nlp models for clinical
  prediction tasks.
\newblock \emph{Journal of biomedical informatics}, 133:104149.

\bibitem[{Otmakhova et~al.(2022)Otmakhova, Verspoor, Baldwin, and
  Lau}]{otmakhova2022patient}
Julia Otmakhova, Karin Verspoor, Timothy Baldwin, and Jey~Han Lau. 2022.
\newblock The patient is more dead than alive: exploring the current state of
  the multi-document summarisation of the biomedical literature.
\newblock In \emph{Proceedings of the 60th Annual Meeting of the Association
  for Computational Linguistics (Volume 1: Long Papers)}, pages 5098--5111.

\bibitem[{Ouyang et~al.(2022)Ouyang, Wu, Jiang, Almeida, Wainwright, Mishkin,
  Zhang, Agarwal, Slama, Ray et~al.}]{ouyang2022training}
Long Ouyang, Jeffrey Wu, Xu~Jiang, Diogo Almeida, Carroll Wainwright, Pamela
  Mishkin, Chong Zhang, Sandhini Agarwal, Katarina Slama, Alex Ray, et~al.
  2022.
\newblock Training language models to follow instructions with human feedback.
\newblock \emph{Advances in Neural Information Processing Systems},
  35:27730--27744.

\bibitem[{Pavao et~al.(2022)Pavao, Guyon, Letournel, Baró, Escalante,
  Escalera, Thomas, and Xu}]{codalab_competitions}
Adrien Pavao, Isabelle Guyon, Anne-Catherine Letournel, Xavier Baró, Hugo
  Escalante, Sergio Escalera, Tyler Thomas, and Zhen Xu. 2022.
\newblock \href {https://hal.inria.fr/hal-03629462v1} {Codalab competitions: An
  open source platform to organize scientific challenges}.
\newblock \emph{Technical report}.

\bibitem[{Radford et~al.(2019)Radford, Wu, Child, Luan, Amodei, Sutskever
  et~al.}]{radford2019language}
Alec Radford, Jeffrey Wu, Rewon Child, David Luan, Dario Amodei, Ilya
  Sutskever, et~al. 2019.
\newblock Language models are unsupervised multitask learners.
\newblock \emph{OpenAI blog}, 1(8):9.

\bibitem[{Raffel et~al.(2020)Raffel, Shazeer, Roberts, Lee, Narang, Matena,
  Zhou, Li, and Liu}]{raffel2020exploring}
Colin Raffel, Noam Shazeer, Adam Roberts, Katherine Lee, Sharan Narang, Michael
  Matena, Yanqi Zhou, Wei Li, and Peter~J Liu. 2020.
\newblock Exploring the limits of transfer learning with a unified text-to-text
  transformer.
\newblock \emph{The Journal of Machine Learning Research}, 21(1):5485--5551.

\bibitem[{Soldaini and Goharian(2016)}]{soldaini2016quickumls}
Luca Soldaini and Nazli Goharian. 2016.
\newblock Quickumls: a fast, unsupervised approach for medical concept
  extraction.
\newblock In \emph{MedIR workshop, sigir}, pages 1--4.

\bibitem[{Taylor et~al.(2022)Taylor, Kardas, Cucurull, Scialom, Hartshorn,
  Saravia, Poulton, Kerkez, and Stojnic}]{taylorgalactica}
Ross Taylor, Marcin Kardas, Guillem Cucurull, Thomas Scialom, Anthony
  Hartshorn, Elvis Saravia, Andrew Poulton, Viktor Kerkez, and Robert Stojnic.
  2022.
\newblock Galactica: A large language model for science.
\newblock \emph{arXiv preprint arXiv:2211.09085}.

\bibitem[{Uzuner et~al.(2011)Uzuner, South, Shen, and DuVall}]{uzuner20112010}
{\"O}zlem Uzuner, Brett~R South, Shuying Shen, and Scott~L DuVall. 2011.
\newblock 2010 i2b2/va challenge on concepts, assertions, and relations in
  clinical text.
\newblock \emph{Journal of the American Medical Informatics Association},
  18(5):552--556.

\bibitem[{Weed(1969)}]{weed1969medical}
Lawrence~L Weed. 1969.
\newblock \emph{Medical records, medical education, and patient care: the
  problem-oriented record as a basic tool}.
\newblock Press of Case Western Reserve University.

\bibitem[{Xue et~al.(2018)Xue, Xu, Rodney~Long, Xue, Antani, Thoma, and
  Huang}]{xue2018multimodal}
Yuan Xue, Tao Xu, L~Rodney~Long, Zhiyun Xue, Sameer Antani, George~R Thoma, and
  Xiaolei Huang. 2018.
\newblock Multimodal recurrent model with attention for automated radiology
  report generation.
\newblock In \emph{Medical Image Computing and Computer Assisted
  Intervention--MICCAI 2018: 21st International Conference, Granada, Spain,
  September 16-20, 2018, Proceedings, Part I}, pages 457--466. Springer.

\bibitem[{Yang and Yu(2020)}]{yang2020generating}
Zhichao Yang and Hong Yu. 2020.
\newblock Generating accurate electronic health assessment from medical graph.
\newblock In \emph{Proceedings of the Conference on Empirical Methods in
  Natural Language Processing. Conference on Empirical Methods in Natural
  Language Processing}, volume 2020, page 3764. NIH Public Access.

\end{thebibliography}
\bibliographystyle{acl_natbib}




\end{document}